\documentclass[11pt, a4paper, logo, copyright]{googledeepmind}

\pdftrailerid{redacted}

\makeatletter
\renewcommand\bibentry[1]{\nocite{#1}{\frenchspacing\@nameuse{BR@r@#1\@extra@b@citeb}}}
\makeatother

\usepackage{kantlipsum, lipsum}
\usepackage{dsfont}
\usepackage{gdm-colors}
\usepackage[page,toc,titletoc,title]{appendix}

\usepackage[authoryear, sort&compress, round]{natbib}

\usepackage{pax}

\title{Generative AI Misuse: A Taxonomy of Tactics and Insights from Real-World Data}

\correspondingauthor{Nahema Marchal <nahemamarchal@google.com>}

\paperurl{arxiv.org/abs/123}

\reportnumber{001} 

\renewcommand{\today}{2024-06-05}

\author[*,1]{Nahema Marchal}
\author[*,2]{Rachel Xu}
\author[3]{Rasmi Elasmar}
\author[1]{Iason Gabriel}
\author[2]{Beth Goldberg}
\author[1]{William Isaac}

\affil[*]{Equal contributions}
\affil[1]{Google DeepMind}
\affil[2]{Jigsaw}
\affil[3]{Google.org}

\begin{abstract}
Generative, multimodal artificial intelligence (GenAI) offers transformative potential across industries, but its misuse poses significant risks. Prior research has shed light on the potential of advanced AI systems to be exploited for malicious purposes. However, we still lack a concrete understanding of how GenAI models are specifically exploited or abused in practice, including the tactics employed to inflict harm. In this paper, we present a taxonomy of GenAI misuse tactics, informed by existing academic literature and a qualitative analysis of approximately 200 observed incidents of misuse reported between January 2023 and March 2024. Through this analysis, we illuminate key and novel patterns in misuse during this time period, including potential motivations, strategies, and how attackers leverage and abuse system capabilities across modalities (e.g. image, text, audio, video) in the wild. 
\end{abstract}

\begin{document}
\renewcommand{\figureautorefname}{Figure}
\renewcommand{\tableautorefname}{Table}
\renewcommand{\appendixautorefname}{Appendix}
\renewcommand{\sectionautorefname}{Section}
\renewcommand{\subsectionautorefname}{Section}
\renewcommand{\subsubsectionautorefname}{Section}
\newcommand{\aref}[1]{\hyperref[#1]{Appendix~\ref*{#1}}}

\maketitle

\clearpage

\tableofcontents

\clearpage

\section*{Acknowledgements}
We would like to thank Mikel Rodriguez, Vikay Bolina, Alexios Mantzarlis, Seliem El-Sayed, Mevan Babakar, Matt Botvinick, Canfer Akbulut, Harry Law, Sébastien Krier, Ziad Reslan, Boxi Wu, Frankie Garcia, and Jennie Brennan, for their feedback and contributions to this paper.

\clearpage

\section{Introduction}
\label{sec:introduction}

Generative, multimodal artificial intelligence (GenAI) brings forth new possibilities across industries and creative domains. Over the past year, leading AI labs have unveiled models that demonstrate sophisticated capabilities across tasks: from complex audiovisual understanding and mathematical reasoning \citep{Gemini_Team2024-ul} to realistic simulation of real-world environments \citep{Brooks2024-tu}. These systems are rapidly being integrated into critical sectors like healthcare \citep{Yim2024-cu}, education \citep{Qadir2023-es} and public services \citep{Bright2024-zs}. Yet, as GenAI capabilities advance, so does awareness of these tools’ potential for misuse, including heightened concerns around security, privacy and manipulation \citep{Feuerriegel2023-eg,Pauly2024-yy,Golda2024-fy,Barrett2023-gv,Shevlane2023-av}.

Prior research has shed light on the potential of advanced AI systems to be exploited for malicious purposes using foresight analysis and hypothetical scenarios, which aim to map out future ethical risks in a systematic way \citep{Brundage2018-pk,Blauth2022-vl, Barrett2023-gv, Goldstein2023-kk, Ferrara2024-bv,Rodriguez2024-oj}. Complementing this research, initiatives such as the OECD AI incidents monitor (AIM), the AI, Algorithmic, and Automation Incidents and Controversies repository (AIAAIC) and the AI Incident Database initiatives actively record AI-related incidents across applications and sort their associated harms. While these efforts provide a rich foundation for mapping AI-enabled threats, they tend to be broad in scope and focus on identifying potential risks and downstream harms. In contrast, we still don’t know enough about how GenAI tools are specifically exploited and abused by different actors, including the tactics employed. As the technology itself becomes more sophisticated and multimodal, better understanding how these manifest in practice and across modalities, is critical.

In this paper, we first present a taxonomy of GenAI misuse tactics, informed by existing academic literature and a qualitative analysis of ~200 media reports of misuse and demonstrations of abuse of GenAI systems published between January 2023 and March 2024). Based on this analysis, we then illuminate key and novel patterns in GenAI misuse during this time period (see \autoref{sec:findings}: Findings), including potential motivations, strategies, and how attackers leverage and abuse system capabilities across modalities (e.g. image, text, audio, video) in an uncontrolled environment.

\noindent We find that:
\begin{enumerate}
	\item \textbf{Manipulation of human likeness} and \textbf{falsification of evidence} underlie the most prevalent tactics in real-world cases of misuse. Most of these were deployed with a discernible intent to influence public opinion, enable scam or fraudulent activities, or to generate profit. 

	\item The majority of reported cases of misuse do not consist of \textbf{technologically sophisticated uses of GenAI systems or attacks}. Instead, we are predominantly seeing an exploitation of easily accessible GenAI capabilities requiring minimal technical expertise.
	\item The increased sophistication, availability and accessibility of GenAI tools seemingly introduces \textbf{new and lower-level forms of misuse} that are neither overtly malicious nor explicitly violate these tools’ terms of services, but still have concerning ethical ramifications. These include the emergence of new forms of communications for political outreach, self-promotion and advocacy that blur the lines between authenticity and deception (see \autoref{sec:discussion}: Discussion).
\end{enumerate}

Our findings provide policy makers, trust and safety teams, and researchers with an evidence base of these technologies’ potential for real-world harm, which can inform their approach to AI governance and mitigations. Second, by providing an overview of salient threats and tactics, this work can also guide the development of safety evaluations and adversarial testing strategies that are more aligned with the rapidly-shifting threat landscape. Lastly, by identifying prominent misuse tactics across modalities this work can help inform targeted mitigations and interventions, with the possibility to better inoculate the public against specific misuse strategies in the future.


\subsection*{Definitions and Scope}
By generative AI (‘GenAI’), we refer to a class of large-scale models trained on billions of parameters of text, audio, code, images and video, mostly sourced from publicly available sources.\footnote{This includes large language and diffusion models such as GPT-4 \citep{OpenAI2023-zj}, Gemini \citep{Gemini_Team2024-ul}, Claude 3 \citep{Anthropic2024-ca}, LLaMA 2 \citep{Touvron2023-sv}, Sora \citep{Brooks2024-tu}, DALL-E 3 \citep{Betker2023-ij}, and Stable Diffusion that can generate outputs across text, image, video, audio and code modalities.} These models, which include large language and diffusion models, can be adapted to a range of downstream tasks, including text and media generation, problem solving \citep{Huang2022-nr}, data extraction \citep{Gartlehner2023-jw}, and coding assistance \citep{Tian2023-nf} among others. They also show several novel capabilities such as the ability to learn how to use external tools \citep{Schick2023-oo} and to perform new tasks without needing to be retrained on large datasets \citep{Wang2023-vr, Wei2022-fa}.

Throughout the paper, building on the definition proposed by \cite{Blauth2022-vl} we refer to GenAI ‘misuse’ as the deliberate use of generative AI tools by individuals and organisations to facilitate, augment or execute actions that may cause downstream harm, as well as attacks on generative AI systems themselves. This definition excludes accidents or cases where harm is caused by malfunctions or limitations of GenAI systems themselves, such as their tendency to hallucinate facts or produce biassed outputs \citep{Maynez2020-xl, Ji2023-rr}, without a discernible actor involved.

\section{Methodology}
\label{sec:methodology}
To develop our taxonomy, we first conducted a review of recent academic and grey literature focusing on malicious uses of generative AI. This initial review provided the initial theoretical foundations for identifying and categorising misuse tactics. We then collected and qualitatively analysed a dataset of media reports of GenAI misuse, as defined above, to validate and augment our taxonomy. Our dataset consists of individual media reports published between January 2023 and March 2024 documenting one or more proof points of real-world misuse involving GenAI. Two authors first independently reviewed each media report in the dataset to identify relevant misuse tactics employed. Our initial taxonomy categories were then continuously updated and expanded based on emerging patterns in the data. Any disagreements between authors were thoroughly discussed to reach consensus, ensuring consistency and accuracy in the classification process.

Whenever possible or clearly identifiable from the reporting, we also extracted information about the actors involved in the misuse, their underlying goals\footnote{While inferring motivations or intent with complete certainty is not always possible for every instance of misuse, for our analysis we rely on contextual information provided from the reporting to offer an educated guess.}, and the individuals organisations or models targeted. This enabled us to identify broader misuse strategies emerging from the combination of specific goals, tactics, uses of GenAI applications, and targets. We present these findings in \autoref{sec:findings} (See \aref{app:goals} and \aref{app:strategies} for a complete list of identified goals and strategies).

To ensure a good coverage of GenAI misuses in our dataset, we employed two data collection approaches.  First, we leveraged a proprietary social listening tool that aggregates content from millions of sources --- including social media platforms such as X and Reddit, blogs and established news outlets --- to detect potential abuses of GenAI tools. We supplemented this with an additional manual search for relevant articles from reputable news sources and blogs published between Jan 1st, 2023 and March 5th, 2024, based on a list of keywords relevant to GenAI misuse.\footnote{Our search strategy included a combination of GenAI-related terms and models (AI-generated, generative AI, GenAI, deepfake*, Gemini, GPT*, Claude, Llama, DALL-E, Midjourney, Stable Diffusion, Bard, Galactica, Sora) and misuse-related terms (harm*, use*, misuse*, abuse*, harass*, victim*, attack*, safety, malicious, manipulat*, dece*, false, fake, content*, data)} Data was then parsed to identify and remove duplicates. After de-deduplication and removal of out-of-scope cases, our dataset contains a total of 191 cases.

While using media reports as a primary source of data allows us to capture emerging trends and novel techniques as they are deployed, and emphasise cases that cause significant harm or disruption, there are nevertheless some limitations to this approach (see \autoref{sec:limitations}: Limitations). Notably, we acknowledge the potential for underreporting of covert misuse operations or instances where the GenAI tactics used are novel or difficult to detect. Additionally, cases that cause less noticeable harm may receive less attention in the press, creating potential blindspots in our data. 

\section{Taxonomy of Generative AI Misuse Tactics}
\label{sec:taxonomy}
To systematically categorise GenAI misuse tactics, we propose a taxonomy that distinguishes between two types of tactics: (1) tactics that involve the exploitation of GenAI’s capabilities, and (2) tactics that involve compromising or attacking GenAI systems themselves.

\subsection{Exploitation of GenAI capabilities}
Across modalities, GenAI models are characterised by their ability to synthesise highly-realistic outputs \citep{Cooke2024-tb,Nightingale2022-gl}, including convincingly mimicking writing and artistic styles \citep{Syed2020-na}. They can produce vast amounts of this content efficiently, allowing for rapid distribution of a high volume of synthetic content in a short time frame. GenAI tools are also characterised by their widespread availability and their ease of use, with intuitive dialogue interfaces that require minimal technical knowledge. While these features enhance the quality and ease of user experiences with GenAI tools, each of these also provide opportunities for exploitation. We identify 10 distinct tactics that exploit GenAI capabilities, summarised in \autoref{tab:tacics_capabilities} below.

\begin{table}[tb]
	\centering
	\caption{Misuse tactics that exploit GenAI capabilities}
	\includegraphics[width=\columnwidth]{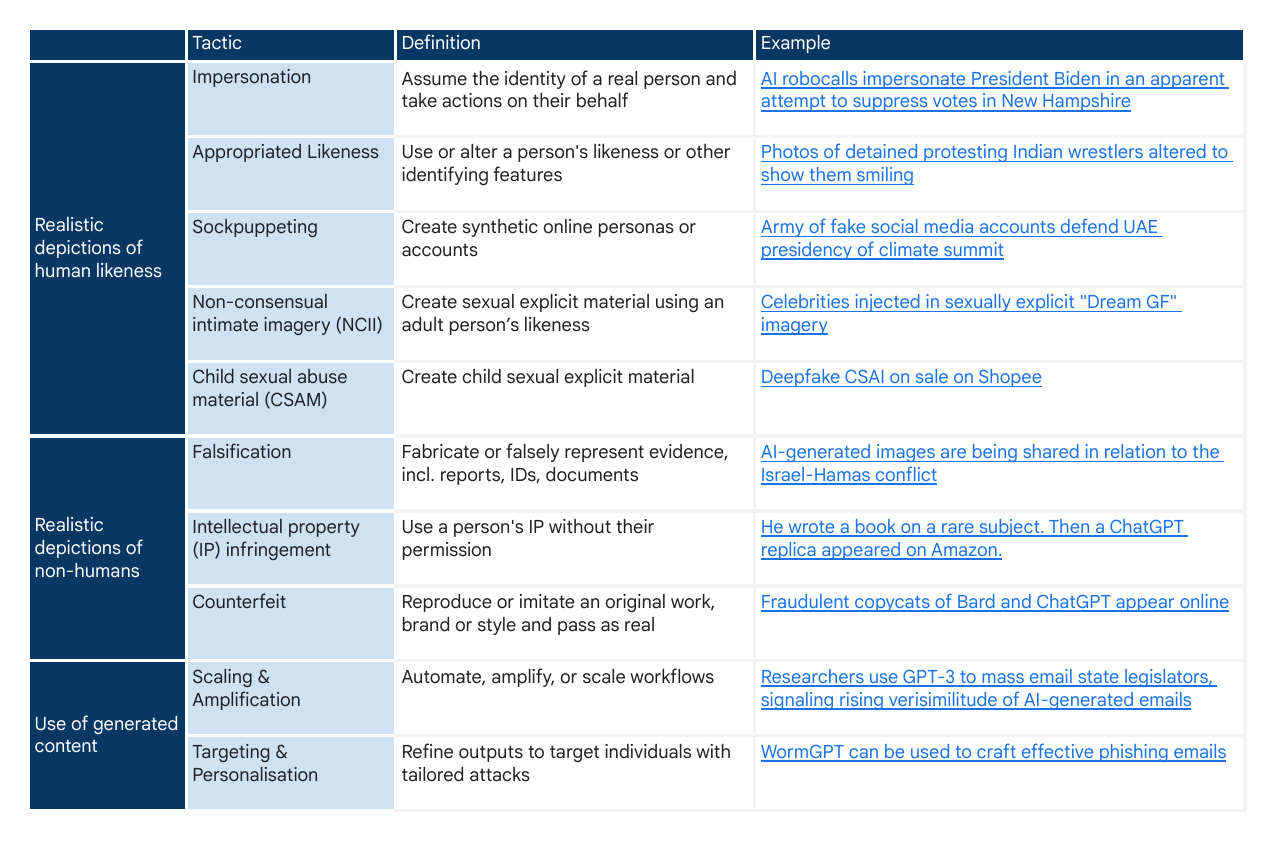}
\label{tab:tacics_capabilities}
\end{table}

The first five tactics listed leverage GenAI models to create realistic depictions of people from natural language descriptions. To distinguish between tactics, we used three framing questions that we believe yield different trust and safety implications: (1) Does the generated output depict a real person or an entirely synthetic one? (2) Is the depiction static or in real-time interaction? and (3) How is the generated content being used?

When the GenAI output depicts a real person, and is ostensibly used to take action on their behalf in real time (e.g. an AI-generated audio clip of someone, in which they claim to be a famous politician), we term this tactic \textbf{Impersonation}. Active depictions such as these carry a high potential for harm, as they can easily mislead an unwitting audience due to their realism and resemblance to everyday experiences and scenarios \citep{Vaccari2020-uo,Sundar2008-qc}. When an output depicts a real person in a static manner and at a fixed point in time (e.g. a synthetic image of a celebrity in a country they’ve never visited), we term this tactic as \textbf{Appropriated Likeness}. This often leverages diffusion models’ in-painting and out-painting capabilities --- the ability to make changes to an existing uploaded image (such as removing details) and to complete missing parts of a photo or image beyond its original \citep{Lugmayr2022-tc}.\footnote{See for e.g. \href{https://helpx.adobe.com/express/using/generative-fill.html}{Insert or replace objects with Generative fill}}  When a GenAI tool is used to depict entirely synthetic personas and make them take action in the world, we call this tactic \textbf{Sockpuppeting}. While not directly targeting specific individuals, the creation of synthetic personas, such as fake experts and activists, can lend a powerful force multiplier to actors seeking to shape and manipulate information in digital environments \citep{Harris2023-na}.\footnote{See for e.g. \href{https://twitter.com/BBCMarcoSilva/status/1690665259298680832}{Marco Silva}}

Finally, we separate out the creation of non-consensual sexually explicit material of adults (\textbf{NCII}) and the production of child sexual abuse material (\textbf{CSAM}) even though they may deploy any of the three tactics above. These tactics warrant separate categorization due to their uniquely damaging potential, and our understanding of how practitioners treat this content in practice. Unlike any of the tactics listed above, using GenAI to create CSAM or NCII is generally considered policy violative, regardless of how that content is used. 

The next three tactics leverage audio, image or video-generation capabilities to create realistic depictions of non-humans, including documents, songs, styles, events, places. This includes \textbf{IP Infringement}, where GenAI is used to produce parts or the entirety of someone's intellectual property --- such as literary and artistic works --- without permission. When GenAI is used to create digital items that mimic an original work or style and falsely represent them as authentic (for example, appropriating a reputable news website’s logo and layout), we refer to this tactic as \textbf{Counterfeit}. When a GenAI output depicts fabricated events, places or objects presented as real, we refer to this tactic as \textbf{Falsification}.

Finally, GenAI models can be used to create high volumes of textual or audio-visual content and facilitate distribution and engagement with that content. For example, operating large networks of fake social media profiles that employ LLMs to generate human-like content, images and audio, as well as engage with each other or other users online \citep{DiResta2024-hs}. We refer to this cluster of tactics as \textbf{Scaling and Amplification}. Actors can also leverage similar capabilities to refine model outputs and target them to specific audiences (\textbf{Targeting and Personalisation}), such as translating content into different languages to tailor to different geographies \citep{Yang2023-kv}. Many of these tactics can be used in combination. For example, malicious actors orchestrating large-scale influence operations often use Sockpuppeting and Scaling and Amplification methods together to create botnets --- fake profiles that appear to be real individuals \citep{Gorwa2020-ye}.

\subsection{Compromise of GenAI systems}

Beyond this, we also identify several tactics aiming to compromise GenAI systems themselves. Unlike the tactics presented above, these tactics do not exploit capabilities but instead vulnerabilities in GenAI systems themselves. While there is already extensive literature documenting the exploitation and abuse of AI systems writ large, attempts to game or manipulate GenAI models are relatively new and rapidly-evolving \citep{Rodriguez2024-oj}. We identify 8 tactics in this vein from our analysis of media reports, most of which map on to research demonstrations of model vulnerabilities and possible compromise pathways, rather than genuine attacks from malicious actors (see \autoref{sec:findings}: Findings). These are summarised in \autoref{tab:tactics_compromise} below. 

We separate these tactics based on the part of the system that the compromise is targeted at. Broadly, we distinguish between attacks on: (1) Model integrity (attacks that manipulate the model itself, its structure, settings or input prompts) and (2) Data integrity (attacks that alter the model’s training data or compromise its security and privacy). 

\textbf{Adversarial Inputs} involve modifying individual input data to cause a model to malfunction. These modifications, which are often imperceptible to humans, exploit how the model makes decisions to produce errors \citep{Wallace2019-tx} and can be applied to text, but also to images, audio, or video (e.g. changing pixels in an image of a panda in a way that causes a model to label it as a gibbon).\footnote{See for e.g. \href{https://openai.com/research/attacking-machine-learning-with-adversarial-examples}{Attacking machine learning with adversarial examples}}

\textbf{Prompt Injections} are a form of Adversarial Input that involve manipulating the text instructions given to a GenAI system \citep{Liu2023-vr}. Prompt Injections exploit loopholes in a model’s architectures that have no separation between system instructions and user data to produce a harmful output \citep{Perez2022-cw}. While researchers may use similar techniques to test the robustness of GenAI models, malicious actors can also leverage them. For example, they might flood a model with manipulative prompts to cause denial-of-service attacks or to bypass an AI detection software.

\begin{table}[tb]
	\centering
	\caption{Misuse tactics to compromise GenAI systems}
	\includegraphics[width=\columnwidth]{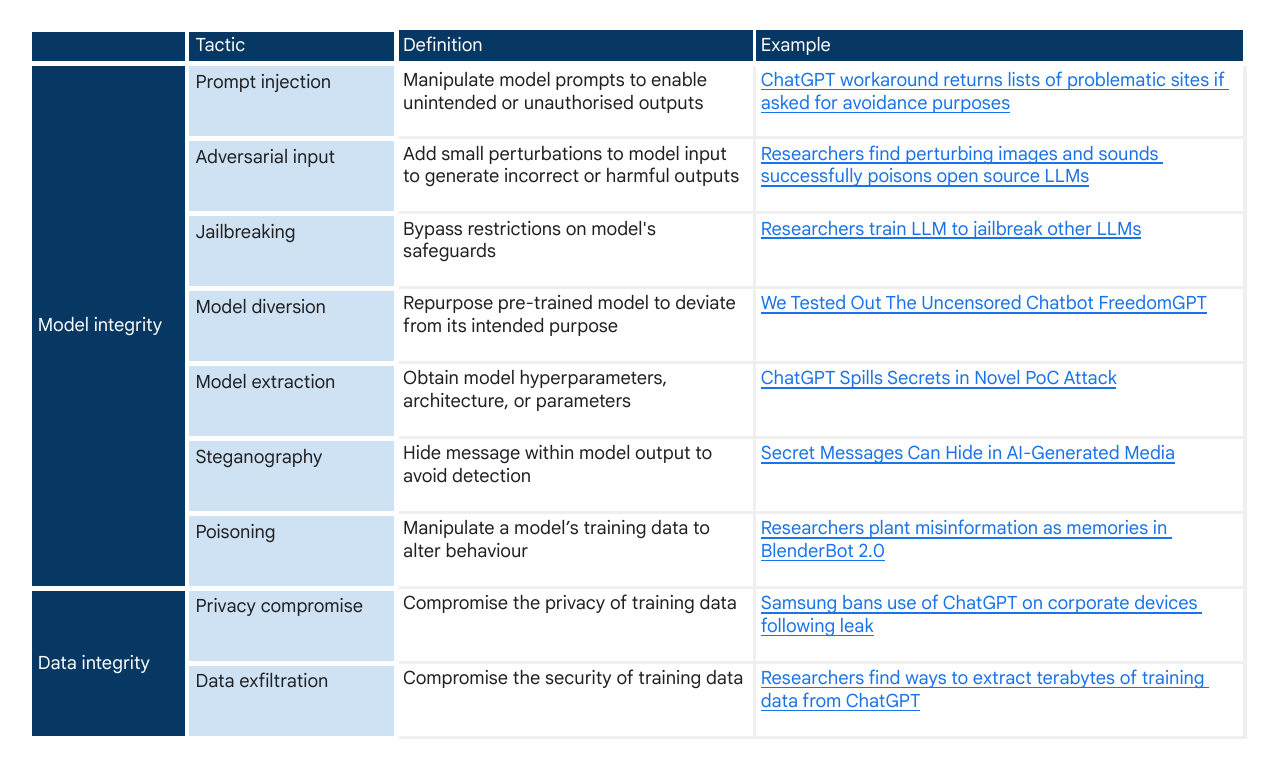}
\label{tab:tactics_compromise}
\end{table}

\textbf{Jailbreaking} aims to bypass or remove restrictions and safety filters placed on a GenAI model completely \citep{Shen2023-mt, Chao2023-dn}. This gives the actor free rein to generate any output, regardless of its content being harmful, biassed, or offensive. All three of these are tactics that manipulate the model into producing harmful outputs against its design. The difference is that prompt injections and adversarial inputs usually seek to steer the model towards producing harmful or incorrect outputs from one query, whereas jailbreaking seeks to dismantle a model's safety mechanisms in their entirety. 

\textbf{Model Diversion} takes model manipulation one step further, by repurposing (often open-source) generative AI models in a way that diverts them from their intended functionality or from the use cases envisioned by their developers \citep{Lin2024-di}. An example of this is training the BERT open source model on the DarkWeb to create DarkBert.\footnote{\href{https://futurism.com/the-byte/ai-trained-dark-web}{Scientists Train New AI Exclusively on the Dark Web}}

\textbf{Steganography} is the practice of hiding coded messages in GenAI model outputs, which may allow malicious actors to communicate covertly.\footnote{\href{https://www.quantamagazine.org/secret-messages-can-hide-in-ai-generated-media-20230518/}{Secret Messages Can Hide in AI-Generated Media}}  \textbf{Data Poisoning} involves deliberately corrupting a model’s training dataset to introduce vulnerabilities, derail its learning process, or cause it to make incorrect predictions \citep{Carlini2023-yv}. For example, the tool Nightshade is a data poisoning tool, which allows artists to add invisible changes to the pixels in their art before uploading online, to break any models that use it for training.\footnote{\href{https://www.technologyreview.com/2023/10/23/1082189/data-poisoning-artists-fight-generative-ai/}{This new data poisoning tool lets artists fight back against generative AI}} Such attacks exploit the fact that most GenAI models are trained on publicly available datasets like images and videos scraped from the web, which malicious actors can easily compromise.

\textbf{Privacy Compromise} attacks reveal sensitive or private information that was used to train a model. For example, personally identifiable information or medical records. \textbf{Data Exfiltration} goes beyond revealing private information, and involves illicitly obtaining the training data used to build a model that may be sensitive or proprietary. \textbf{Model Extraction} is the same attack, only directed at the model instead of the training data --- it involves obtaining the architecture, parameters, or hyper-parameters of a proprietary model \citep{Carlini2024-zu}.

\section{Findings}
\label{sec:findings}
In this section, we summarise findings from our analysisof media reports of GenAI misuse between January 2023and March 2024 to provide an empirically-grounded under-standing of how the threat landscape of GenAI is evolving. Beyond identifying salient misuse tactics, whenever possible and discernible from the reporting we also extracted information for each case in our dataset about the actors involved in the misuse, their goals, and which specific GenAI tools were exploited across modalities. Our analysis reveals patterns in how these elements combine into broader misuse strategies. We tabulate this information in \aref{app:goals} and \aref{app:strategies} and use these data points to enrich our discussion.

\subsection{Prevalence and modalities of misuse tactics}

\begin{figure}[ht]

	\includegraphics[width=\columnwidth]{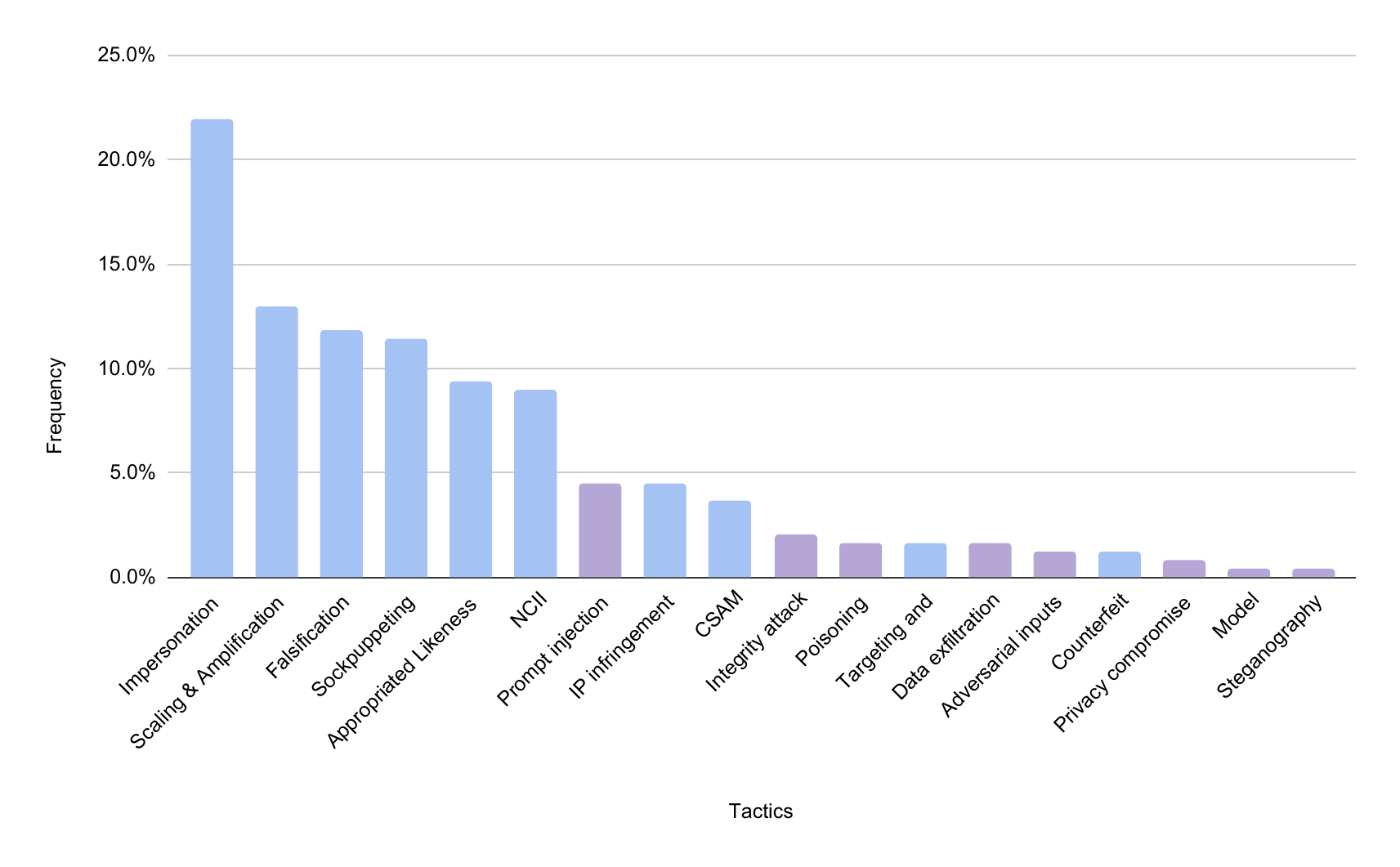}
	\caption{Frequency of tactics across categories.}
	\footnotesize{Note: Each bar represents the frequency with which a tactic was identified within our dataset. Each case of misuse could involve more than one tactic.}
	\label{fig:tactics_frequency}
\end{figure}

First, we find that most reported cases of GenAI misuse involve actors exploiting the capabilities of these systems, rather than launching direct attacks at the models themselves (see \autoref{fig:tactics_frequency}). Nearly 9 out of 10 documented cases in our dataset fall into this category. Of these, the most prevalent cluster of tactics involve the manipulation of human likeness, especially Impersonation (followed by Sockpuppeting, Appropriated Likeness and NCII). Scaling \& Amplification and Falsification are also prominent tactics, accounting for 13\% and 12\% of reported cases respectively.

As \autoref{tab:modalities_tactics} below shows, Impersonation typically involves text-to-speech and video generation tools to replicate people’s voices and likeness, especially that of public figures.\footnote{See for e.g. \href{https://www.politico.com/newsletters/new-york-playbook/2024/01/23/faked-ai-audio-hits-harlem-politics-00137132}{Faked AI audio hits Harlem politics}}  Falsification of content, on the other hand, mostly draws on text and image generation to create synthetic books, news articles and website copy, and images to accompany these articles, such as synthetic images of events that never took place (for e.g., the fake images of explosions at the Pentagon\footnote{\href{https://apnews.com/article/pentagon-explosion-misinformation-stock-market-ai-96f534c790872fde67012ee81b5ed6a4}{Fake image of Pentagon explosion briefly sends jitters through stock market}}). NCII primarily relies on manipulating image and video modalities to create suggestive deepfakes of private individuals or celebrities. Sockpuppeting and tactics centred on scaling and amplifying content distribution involve creating synthetic social media profiles with AI-generated profile pictures and descriptions.

\begin{table}[tb]
	\centering
	\begin{center}
	\caption{Modalities associated with each tactic. }
	\includegraphics[width=4in]{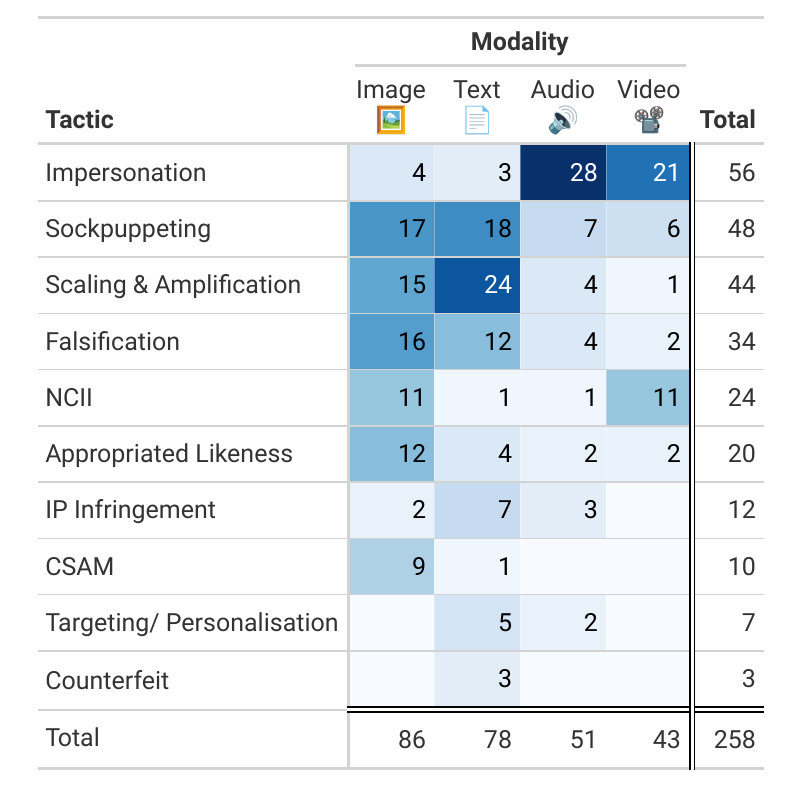}
\label{tab:modalities_tactics}
	\end{center}
\footnotesize{Note: Counts denote the number of times a tactic was linked with this specific modality in our data.}
\end{table}

\subsection{Goals and strategies of misuse}
GenAI misuse does not happen in a vacuum. Actors often have discernible goals or specific motivations to misuse and abuse GenAI, not all of which are necessarily adversarial. These range from financial gain to harassment, and political disruption (see \aref{app:goals} for full breakdown). Understanding these motivations is crucial for assessing the severity of downstream impacts and crafting appropriate countermeasures. 
By observing how actors combine misuse tactics in pursuit of their goals, we can also identify specific patterns of misuse: we label these combinations as strategies (see \aref{app:strategies}) Below, we identify the most common goals behind GenAI misuse, along with the most prevalent and novel strategies used to achieve them.

\begin{table}[h]
	\centering
	\caption{Count of tactics employed per goal}
	\includegraphics[width=\columnwidth]{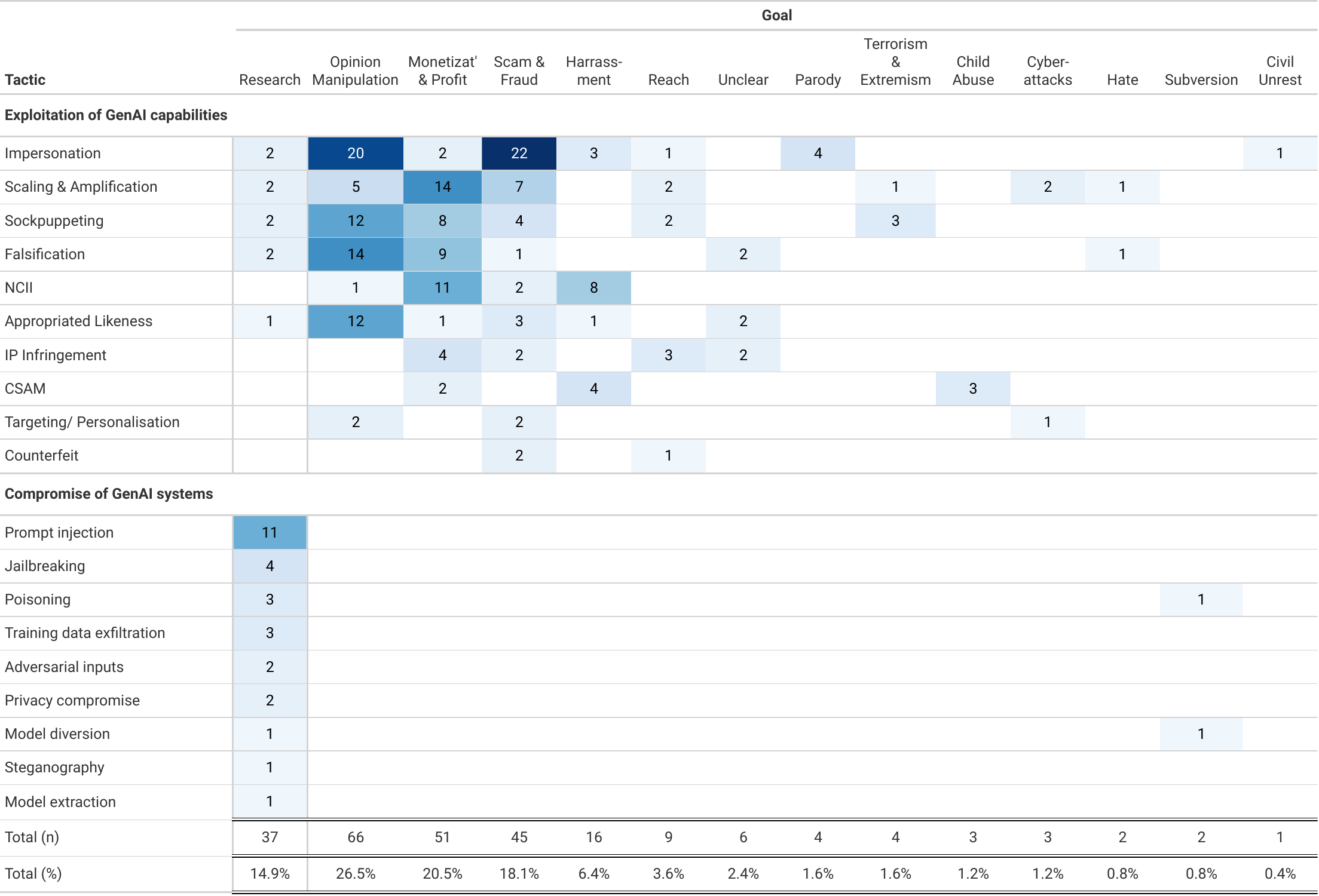}
\label{tab:goals_tactics}
\end{table}

Between 2023-2024, our data shows that \textbf{attacks on GenAI systems themselves were mostly conducted as part of research demonstrations or testing} aimed at uncovering vulnerabilities and weaknesses within these systems (See \autoref{tab:goals_tactics} above). Within this subset, approximately a third of these attempts employed Prompt Injection as a tactic. In contrast, we find limited evidence of attacks on deployed GenAI systems in the wild. Specifically, we document only two real-world instances of compromise, the goals of which were to prevent the unauthorised scraping of copyrighted materials,\footnote{\href{https://www.technologyreview.com/2023/10/23/1082189/data-poisoning-artists-fight-generative-ai/}{This new data poisoning tool lets artists fight back against generative AI}} and provide users with the ability to generate uncensored content.\footnote{\href{https://www.buzzfeednews.com/article/pranavdixit/freedomgpt-ai-chatbot-test}{We Tested Out The Uncensored Chatbot FreedomGPT}} While we find limited evidence of targeted attacks reported in the news, it's important to note that jailbreaking and other model attacks might be occurring, often without widespread publicity. Therefore, the actual number of real-world compromises may be higher than the two instances documented here.

\subsubsection{Opinion Manipulation}
\label{sec:opinion_manipulation}

The most common goal for exploiting GenAI capabilities during this period was \textbf{to shape or influence public opinion} (27\% of all reported cases). In those instances, we saw actors deploy a range of tactics to distort the public’s perception of political realities. These included impersonating public figures, using synthetic digital personas to simulate grassroots support for or against a cause (‘astroturfing’) and creating falsified media.

The majority of cases in our dataset involved the generation of emotionally charged synthetic images around politically divisive topics, such as war, societal unrest or economic decline. For example, images and ads shared during electoral campaigns in the US, Canada and New Zealand by party staffers\footnote{See for e.g. \href{https://www.foxnews.com/politics/house-gop-campaign-arm-slams-democrats-new-ai-generated-ad-turning-national-parks-migrant-tent-cities}{House GOP campaign arm slams Democrats in new AI-generated ad turning national parks into migrant tent cities}; \href{https://www.nytimes.com/2023/06/25/technology/ai-elections-disinformation-guardrails.html}{A.I.’s Use in Elections Sets Off a Scramble for Guardrails}} and state-sponsored actors\footnote{\href{https://www.nytimes.com/2023/09/11/us/politics/china-disinformation-ai.html}{China Sows Disinformation About Hawaii Fires Using New Techniques}; \href{https://www.nytimes.com/2024/02/15/business/media/chinese-influence-campaign-division-elections.html}{Chinese Influence Campaign Pushes Disunity Before U.S. Election, Study Says}} alike frequently depicted scenes of urban decay, homelessness and insecurity. Purportedly ‘leaked’ AI-generated videos and audio clips of politicians falsely endorsing controversial political positions --- such as Vladimir Putin declaring martial law after Ukrainian forces entered Russian territory\footnote{\href{https://www.politico.eu/article/fake-vladimir-putin-announces-russia-under-attack-ukraine-war/}{‘Fake Putin’ announces Russia under attack as Ukraine goes on offensive}} --- and privately attacking their political opponents were also common.\footnote{See for e.g. \href{https://www.bloomberg.com/news/articles/2023-09-29/trolls-in-slovakian-election-tap-ai-deepfakes-to-spread-disinfo}{Trolls in Slovakian Election Tap AI Deepfakes to Spread Disinfo}} These uses of GenAI are clustered under the ‘Disinformation’ label in \autoref{fig:goals_strategies} below.

Defamation was another central strategy for opinion manipulation, with GenAI tools frequently used to impersonate political figures or dissidents and portray them in compromising situations that undermine their reputation or public standing. Specific instances from our data involved depicting electoral candidates spouting abuse towards protected groups, party staffers, or their own constituents.\footnote{See for e.g. \href{https://news.sky.com/story/labour-faces-political-attack-after-deepfake-audio-is-posted-of-sir-keir-starmer-12980181}{Deepfake audio of Sir Keir Starmer released on first day of Labour conference}; \href{https://www.reuters.com/world/us/deepfaking-it-americas-2024-election-collides-with-ai-boom-2023-05-30/}{Deepfaking it: America's 2024 election collides with AI boom}} In other cases, actors shared AI-generated images of politicians appearing visibly aged to make them look unfit for leadership,\footnote{\href{https://www.sochfactcheck.com/viral-video-clip-of-imran-khan-looking-older-is-ai-generated/}{Viral video clip of Imran Khan looking older is AI-generated}} and showing them in intimate settings with other public figures.\footnote{\href{https://factcheck.afp.com/doc.afp.com.33H928Z}{Ron DeSantis ad uses AI-generated photos of Trump, Fauci}}

\begin{figure}[t]
	\centering
	\includegraphics[width=6.1in]{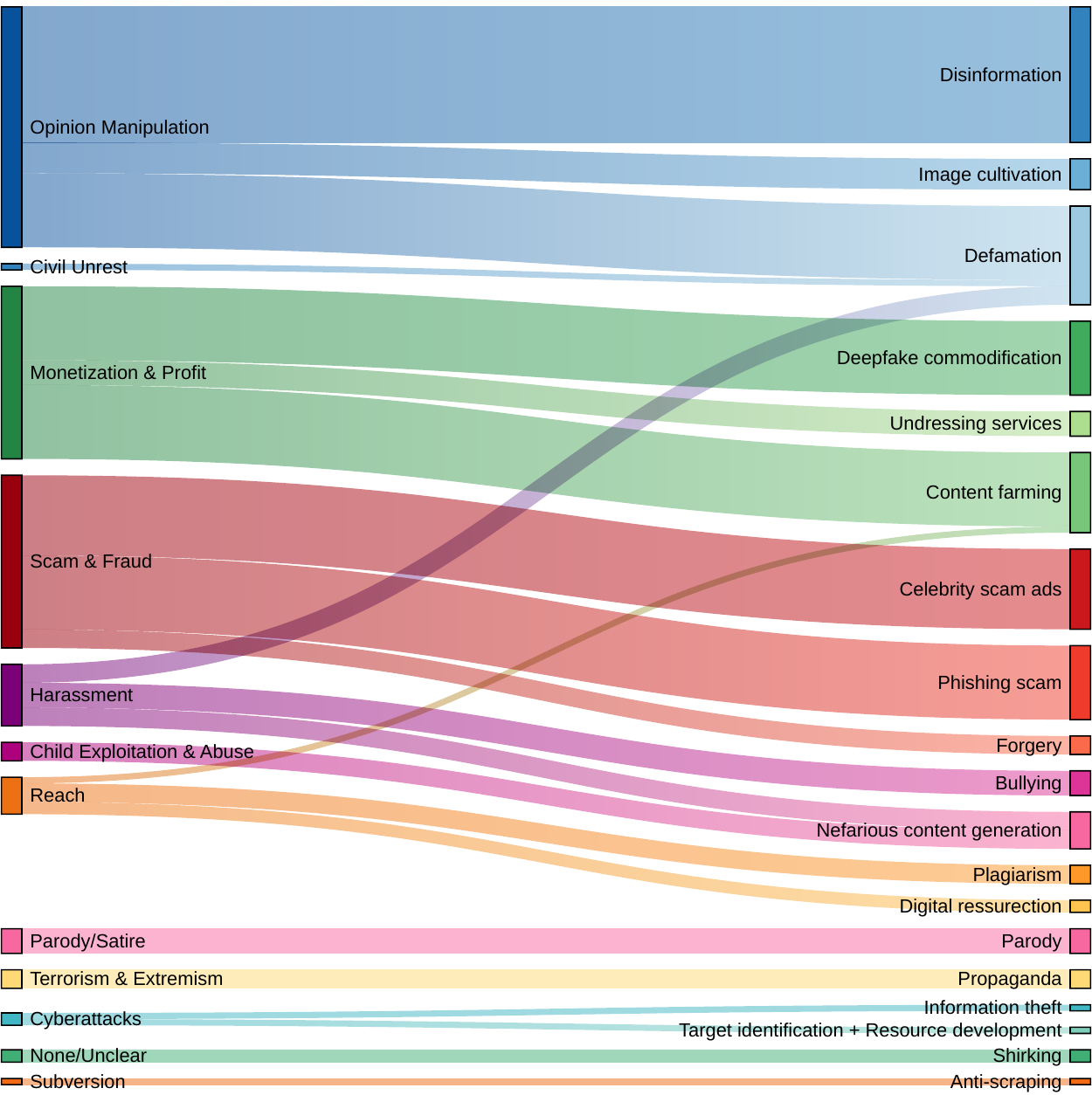}
	\caption{Top strategies associated with each misuse goal.}
	\label{fig:goals_strategies}
\end{figure}

An emerging, though less prevalent, trend was the undisclosed use of AI-generated media by political candidates and their supporters to construct a positive public image. In one case, for example, consultants hired by the team of a Philadelphia sheriff used GenAI to fabricate positive news stories for her campaign website,\footnote{\href{https://www.usnews.com/news/best-states/pennsylvania/articles/2024-02-05/philly-sheriffs-campaign-takes-down-bogus-news-stories-posted-to-site-that-were-generated-by-ai}{Philly Sheriff's Campaign Takes Down Bogus 'News' Stories Posted to Site That Were Generated by AI}} while the campaign staff of one of Argentina’s presidential candidates leveraged GenAI to bolster his image as a strong and charismatic leader, through immersive videos and altered images. These portrayals included depicting him as a soldier in battle and emulating visual styles of Soviet-era propaganda.\footnote{\href{https://www.nytimes.com/2023/11/15/world/americas/argentina-election-ai-milei-massa.html}{Is Argentina the First A.I. Election?}}  Finally,  a handful of cases involved political actors using GenAI for hypertargeted political outreach, such as  simulating their own voice with high fidelity to reach out to their constituents in their native languages, or deploy GenAI-powered campaign callers to engage in tailored conversations with voters on key issues.\footnote{\href{https://www.reuters.com/technology/meet-ashley-worlds-first-ai-powered-political-campaign-caller-2023-12-12/}{Meet Ashley, the world's first AI-powered political campaign caller}} In all of these cases, the lack of appropriate disclosure around the use of GenAI tools in the context of campaigning risks misleading users and causing harm through deception.

\subsubsection{Monetization \& Profit}
The second most common goal behind GenAI misuse was \textbf{to monetize products and services} (21\% of reported cases, see \autoref{tab:goals_tactics}). Profit-driven actors tended to leverage a wide array of tactics, including content scaling, amplification and falsification.

Content farming --- the generation of vast quantities of content at scale --- was a prevalent strategy observed in relation to this goal. This strategy primarily involved private users and, at times, small corporations churning out low quality AI-generated articles, books and product ads for placement on websites such as Amazon and Etsy to cut costs and capitalise on advertising revenue.\footnote{See for e.g. \href{https://www.nytimes.com/2023/08/05/travel/amazon-guidebooks-artificial-intelligence.html}{A New Frontier for Travel Scammers: A.I.-Generated Guidebooks}; \href{https://www.newsguardtech.com/special-reports/newsbots-ai-generated-news-websites-proliferating/}{Rise of the Newsbots: AI-Generated News Websites Proliferating Online}} Aside from this profit-making motive, it is important to note that content farms are also commonly used by state-sponsored actors to flood the information space with false or misleading information.\footnote{\href{https://go.recordedfuture.com/hubfs/reports/cta-2020-0429.pdf}{Chinese Influence Operations Evolve in Campaigns Targeting Taiwanese Elections, Hong Kong Protests}}

The creation of non-consensual intimate imagery (NCII) also represented a significant portion of monetization-driven misuse. In nearly all of these misuse cases, image- and video-generation tools were used to create and sell sexually explicit videos of celebrities who did not consent to the production of that content, or to “nudify” them as a paid service (‘undressing services’).\footnote{See for e.g. \href{https://time.com/6344068/nudify-apps-undress-photos-women-artificial-intelligence/}{‘Nudify’ Apps That Use AI to ‘Undress’ Women in Photos Are Soaring in Popularity}}

\subsubsection{Scam \& Fraud}

Third, actors leveraged GenAI to engage in \textbf{scams and fraudulent activities} such as stealing information, money or other assets (18\%). Fraud-motivated misuses tended to leverage the power of real identities to deceive victims. While identity-based fraud is a long-standing issue, the photorealism and sophistication of GenAI outputs has empowered malicious actors to create more personalised and highly persuasive scams.

Celebrity scam ads, for example, which involve impersonating influential figures to promote fraudulent crypto and investment schemes, were prominent in our dataset. Another common strategy involved using AI-generated audio or video to impersonate trusted individuals, such as a loved one or senior colleague, in a bid to extort money from victims (‘Phishing scams’ in \autoref{fig:goals_strategies}). We also observed the use of GenAI to create bespoke business email compromise campaigns or to convincingly imitate an organisation’s trademark or logo to boost the believability of phishing emails.\footnote{\href{https://cybernews.com/news/generative-ai-fraudulent-emails/}{Generative AI making it harder to spot fraudulent emails}}

These misuse tactics not only infringe upon the targeted individual or organisation’s rights and reputation, but also inflict a potentially high financial and psychological cost to victims. One of the cases in our dataset, for example, saw a financial worker being tricked into transferring \$25m to scammers, who had used GenAI to impersonate several of the employee’s co-workers on  a video call.\footnote{\href{https://edition.cnn.com/2024/02/04/asia/deepfake-cfo-scam-hong-kong-intl-hnk/index.html}{Finance worker pays out \$25 million after video call with deepfake ‘chief financial officer’}}

\subsubsection{Harassment}
Fourth, approximately 6\% of observed cases of GenAI misuse involved some form of harassment or intimidation, with a distinctly gendered dimension. Here, most cases in our dataset centred on the non-consensual generation of NCII (non-consensual intimate imagery) targeting both adults and adolescents, all of which were female. A prevalent and disturbing trend in that respect was the creation and sharing of AI-generated nudes of high school students as part of bullying campaigns. Journalists and celebrities were also common targets of this type of abuse.

Beyond the defamation cases discussed in \autoref{sec:opinion_manipulation} (‘Opinion Manipulation’), outside the political domain, we also noted several instances involving the unauthorised cloning of public figures’ voices and likeness for abuse (including high school principals and celebrities) or as an attempt to defame them.\footnote{See for e.g. \href{https://www.vice.com/en/article/dy7mww/ai-voice-firm-4chan-celebrity-voices-emma-watson-joe-rogan-elevenlabs}{AI-Generated Voice Firm Clamps Down After 4chan Makes Celebrity Voices for Abuse}; \href{https://www.vice.com/en/article/7kxzk9/school-principal-deepfake-racist-video}{High Schoolers Made a Racist Deepfake of a Principal Threatening Black Students}} A rarely observed but seemingly novel form of harassment involved malicious actors generating audio clips of voice actors doxxing themselves (i.e. reading aloud their own addresses) and sharing them online. This case, and others mentioned above, highlight a growing and concerning risk of targeted abuse for content creators and individuals whose data is easily or publicly accessible.

\subsubsection{Reach}
Finally, a comparatively small proportion of cases involved using GenAI to maximise the reach of a message, piece of content, or brand (3.6\%). Though still marginal, one noteworthy and emerging strategy in relation to this goal was the rise of digital resurrections for reach and advocacy. We saw multiple cases of GenAI being used to recreate the likeness of individuals who had passed away, sometimes without the consent of those involved. In Feb 2024, two activist groups used GenAI to create voice recordings of school shooting victims in an attempt to compel Congress to act on gun violence.\footnote{\href{https://www.theguardian.com/us-news/2024/feb/14/ai-shooting-victims-calls-gun-reform}{Voices of the dead: shooting victims plead for gun reform with AI-voice messages}} In Aug 2023, content creators on TikTok used GenAI to “give voices” to deceased or missing children to narrate disturbing details of their experiences in an attempt to “raise awareness.”\footnote{\href{https://www.washingtonpost.com/technology/2023/08/09/ai-dead-children-tiktok-videos/}{AI is being used to give dead, missing kids a voice they didn’t ask for}}

While not necessarily violative of a model’s content policies, practices such as these raise profound ethical concerns by instrumentalizing the likeness of people who cannot express consent or lack agency over how their image is utilised.

\section{Discussion}
\label{sec:discussion}

Our analysis of real-world GenAI misuse highlights key patterns with significant implications for trust and safety practitioners, policy makers and researchers.

Our data shows that GenAI tools are primarily exploited to \textbf{manipulate human likeness} (through Impersonation, Sockpuppeting, Appropriated Likeness and NCII) and \textbf{falsify evidence}. The prevalence of these tactics may be due to the fact that sources of human data (e.g. images, audio, video) abound online, making it relatively easy for bad-faith actors to feed this information into generative AI systems. However, it is also possible that these types of misuses simply attract more media attention than others, due to their broad societal impact. These cases of misuse primarily aimed to shape public opinion, especially through defamation and manipulation of political perceptions, and to facilitate scams, fraud and quick monetization schemes.

Despite widespread concerns around highly sophisticated, state-sponsored uses of GenAI\footnote{See for e.g. \href{https://assets.publishing.service.gov.uk/media/653932db80884d0013f71b15/generative-ai-safety-security-risks-2025-annex-b.pdf}{Safety and Security Risks of Generative Artificial Intelligence to 2025}; \href{https://www.brookings.edu/articles/propaganda-foreign-interference-and-generative-ai/}{Propaganda, foreign interference, and generative AI}}, we find that \textbf{most cases of GenAI misuse are not sophisticated  attacks on AI systems} but readily exploit easily accessible GenAI capabilities that require minimal technical expertise. Many of the salient tactics we observe, such as impersonation scams, forgery, and synthetic personas, pre-date the invention of GenAI and have long been used to influence the information ecosystem and manipulate others. However, by giving these age-old tactics new potency and democratising access, \textbf{GenAI has altered the costs and incentives associated with information manipulation}, leading to a wide range of use cases and a wide pool of individuals being involved in these activities. As our data shows, this includes political figures and private citizens alike, and those without significant technical background.

The widespread availability, accessibility and hyperrealism of GenAI outputs across modalities has also enabled \textbf{new, lower-level forms of misuse that blur the lines between authentic presentation and deception}. While these uses of GenAI --- such as generating and repurposing content at scale and leveraging GenAI for personalised political communication --- are often \textbf{neither overtly malicious nor explicitly violate these tools’ content policies or terms of services}, their potential for harm is significant. GenAI-powered political image cultivation and advocacy without appropriate disclosure, for example, undermines public trust by making it difficult to distinguish between genuine and manufactured portrayals. Likewise, the mass production of low quality, spam-like and nefarious synthetic content\footnote{See for e.g. \href{https://www.myfloridalegal.com/sites/default/files/2023-09/54-state-ags-urge-study-of-ai-and-harmful-impacts-on-children.pdf}{Re: Artificial Intelligence and the Exploitation of Children}} risks increasing people’s scepticism towards digital information altogether and overloading users with verification tasks. If unaddressed, this contamination of publicly accessible data with AI-generated content could potentially impede information retrieval and distort collective understanding of socio-political reality or scientific consensus. For example, we are already seeing cases of liar’s dividend,\footnote{See for e.g. \href{https://restofworld.org/2023/indian-politician-leaked-audio-ai-deepfake/}{An Indian politician says scandalous audio clips are AI deepfakes. We had them tested}} where high profile individuals are able to explain away unfavourable evidence as AI-generated, shifting the burden of proof in costly and inefficient ways.

These findings carry several consequences for how we approach mitigations. Common misuse tactics such as NCII are exacerbated by technical vulnerabilities within GenAI systems that model developers are actively working to resolve and create safeguards against, such as removing toxic content from training data or restricting prompts that violate these tools’ terms of services. However, many of the cases identified (e.g. those involving deceptive portrayals) prey on vulnerabilities in the broader social context in which they are deployed --- for example, phishing scams campaigns that rely on an individuals’ reasonable expectation of the authenticity of their digital landscape and their interactions with it. While technical interventions may provide some benefit, in these cases, non-technical, user-facing interventions are necessary. Prebunking, for example --- a common psychological intervention to protect against information manipulation \citep{Roozenbeek2022-vp} --- could be usefully extended to protect users against GenAI-enabled deceptive and manipulative tactics.

Additionally, many prevalent forms of misuse hinge on exploiting GenAI capabilities that model developers are actively working to enhance (for example, photo-realistic outputs). As GenAI tools become more capable and accessible, we may therefore continue to see an increase in AI-generated content as part of media-based misinformation and manipulation campaigns \citep{dufour2024ammeba}. While several solutions like synthetic media detection tools and watermarking techniques have been proposed and offer promise, they are far from panaceas \citep{Sadasivan2023-mv}. Notably, the inherent adaptability of malicious actors means that as detection methods improve, so will methods of circumvention \citep*{Leibowicz2021-va}. In these cases, targeted interventions such as restrictions on specific model capabilities and usage restrictions may be warranted when the risk for misuse is high and other interventions are insufficient \citep{Anderljung2023-ba, Shevlane2022-lk}.

\section{Limitations and further research}
\label{sec:limitations}
While this research provides valuable insights into the current landscape of GenAI misuse, it is important to acknowledge several limitations that may affect the generalisability and comprehensiveness of our findings.

First, relying on media reports as a primary data source can introduce biases. Media outlets often prioritise incidents with sensational elements or those that directly impact human perception, potentially skewing our dataset towards particular types of misuse. Conversely, covert attacks or those that do not necessarily include humans in-the-loop --- such as the use of GenAI to obfuscate malicious code to evade filters --- may be underrepresented in our data due to limited media coverage or to the fact that companies may keep this information private. This underscores the need for better and more comprehensive sources of anonymized data --- akin to the Safety Information Analysis and Sharing (ASIAS) System for the aviation industry for example --- to gain a more holistic understanding of the threat landscape and inform effective mitigations.

Second, our analysis is time-bound and therefore only offers a snapshot of GenAI misuse at a specific point in time. Yet, as GenAI models continue to acquire new capabilities, become more agentic and integrated into everyday applications and services, their potential for misuse may expand beyond the current scope of our findings. For example, the progressive integration of GenAI into social media platforms to deliver personalised content may lead to new forms of information manipulation. Keeping up with this dynamic landscape calls for further longitudinal analyses and continued monitoring of emerging tactics and harms. 

Finally, we note that the majority of observed cases of misuse in our dataset involve models that take text prompts as input rather than leveraging truly multimodal capabilities. This is possibly due to the fact that we are analysing data at an early stage in this technology’s development. However, the field is rapidly progressing towards more sophisticated multimodal models capable of processing and generating diverse forms of content. We anticipate that new modalities and capabilities, such as the ability to prompt models with video and image input will likely lead to new patterns of misuse, which should be thoroughly investigated. 

\section{Conclusion}
\label{sec:conclusion}
This research has sought to illuminate the evolving landscape of GenAI misuse, and its impacts. While fears of sophisticated adversarial attacks have dominated public discourse, our findings reveal a prevalence of low-tech, easily accessible misuses by a broad range of actors, often driven by financial or reputational gain. These misuses, while not always overtly malicious, have far-reaching consequences for trust, authenticity, and the integrity of information ecosystems. We have also seen how GenAI amplifies existing threats by lowering barriers to entry and increasing the potency and accessibility of previously costly tactics. These findings underscore the need for a multi-faceted approach to mitigating GenAI misuse, involving collaboration between policymakers, researchers, industry leaders, and civil society. Addressing this challenge requires not only technical advancements but also a deeper understanding of the social and psychological factors that contribute to the misuse of these powerful tools.

\clearpage

\bibliographystyle{plainnat}
\nobibliography*
\bibliography{references}


%

\clearpage
\appendix
\appendixpage

\setcounter{table}{0}
\renewcommand{\thetable}{\Alph{section}.\arabic{table}}

\section{Goals}
\label{app:goals}
Analysing our dataset, we have identified 16 distinct goals driving GenAI misuse (see \autoref{tab:appendix_goals} below).

\begin{table}[h]
	\centering
	\caption{Goals of GenAI misuse}
	\includegraphics[width=\columnwidth]{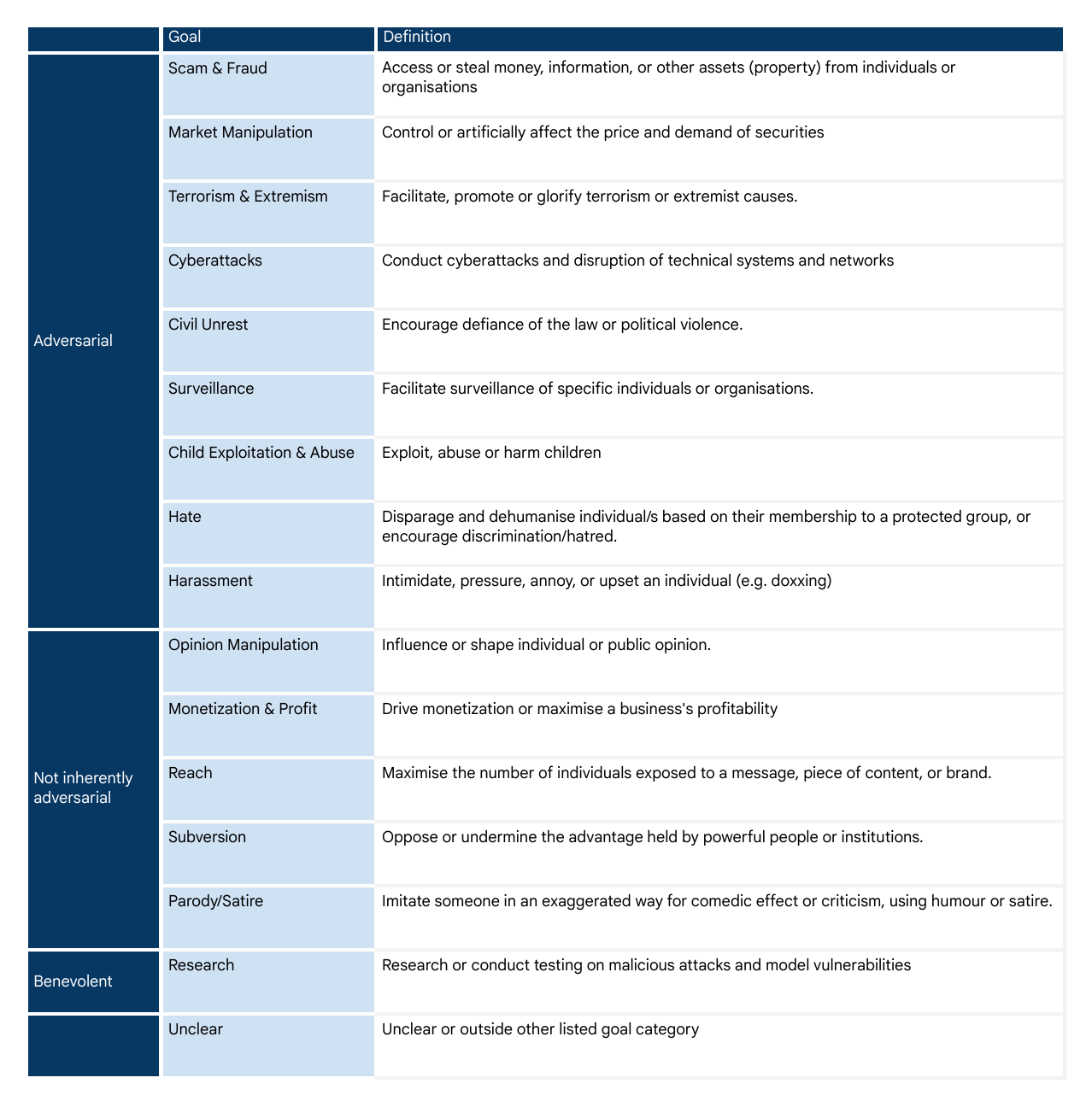}
\label{tab:appendix_goals}
\end{table}

Importantly, not all goals associated with misuse are necessarily adversarial or involve malicious intent. Some goals are, by definition, malicious in that they involve engaging in explicitly illegal activities or demonstrate a clear intent to cause harm. This includes leveraging GenAI models to engage in fraudulent activities such as stealing information, money or other assets (\textbf{Scam \& Fraud}), encouraging political violence and defiance of the law (\textbf{Civil Unrest}), spying on individuals’ online activities and data (\textbf{Surveillance}), conducting cyberattacks, or promoting terrorism and other extreme ideologies (\textbf{Terrorism \& Extremism}). Other goals that fit in this category include exploiting GenAI tools to attack, dehumanise or disparage protected groups (\textbf{Hate}), harass individuals (\textbf{Harassment}) and exploit or harm children (\textbf{Child Exploitation and Abuse}). 

In contrast, some goals may be pursued without an explicit intention to cause harm, but could still result in adverse consequences for individuals and society. This includes cases where actors use GenAI to try to influence public opinion about socio-political issues (\textbf{Opinion Manipulation}), to sell products and services or to maximise a business profitability (\textbf{Monetization \& Profit}), or to manipulate the price and demand of stocks and securities (\textbf{Market Manipulation}). Other relevant goals include leveraging GenAI to aid the distribution of specific messages or content (\textbf{Reach}), to oppose or undermine power structures (\textbf{Subversion}), or for satirical purposes (\textbf{Parody}).

Finally, we observe several instances where generative AI is only misused for ostensibly benevolent purposes, as part of research efforts and testing aimed at exposing model vulnerabilities (\textbf{Research}). Of course, it is possible for any given case of misuse to have more than one goal at a time. For simplicity and clarity, in our dataset, we track only what we believed to be the primary goal of each case of misuse, based on the contextual information provided by media reports.

\section{Strategies}
\label{app:strategies}

Our observations also reveal distinct combinations of goals, tactics, uses of GenAI and targets of misuse coalescing into broader 'misuse strategies.' These strategies are useful to delineate as they reveal the calculated steps taken to leverage GenAI towards different ends, which may demand tailored interventions or mitigation strategies. In the following table, we enumerate these strategies, organised by goal, and outline the tactics they employ, along with salient examples from our dataset.

\setcounter{table}{0}

\clearpage

\begin{table}[ht]
	\centering
	\caption{Strategies of GenAI misuse per goal}
	\includegraphics[width=\columnwidth]{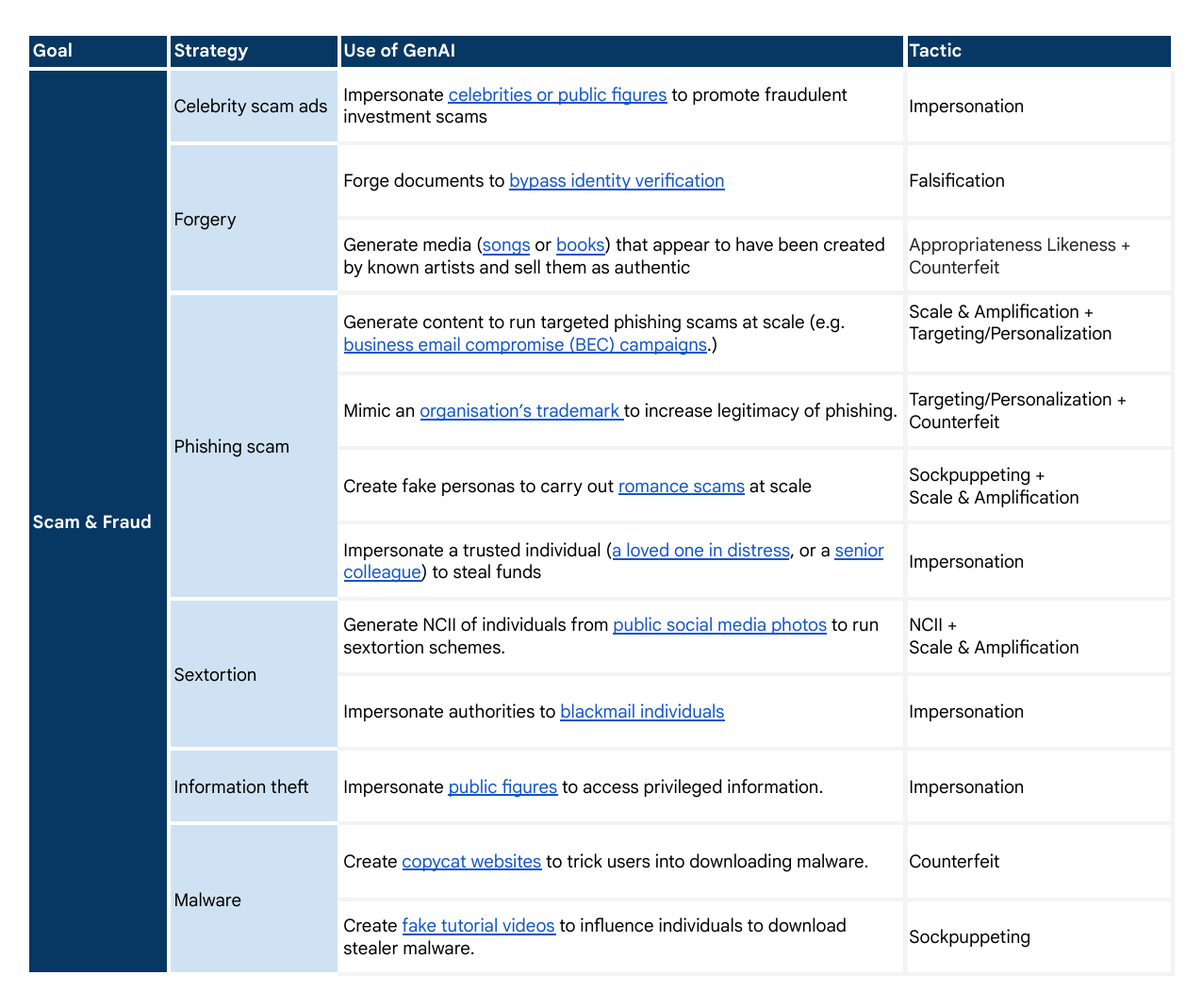}
	\captionsetup{justification=centering}
	\caption*{\textit{(continued on the next page)}}
\label{tab:appendix_strategies}
\end{table}

\clearpage
\setcounter{table}{0}
\begin{table}[ht]
	\centering
	\caption{Strategies of GenAI misuse per goal \textit{(continued)}}
	\includegraphics[width=\columnwidth]{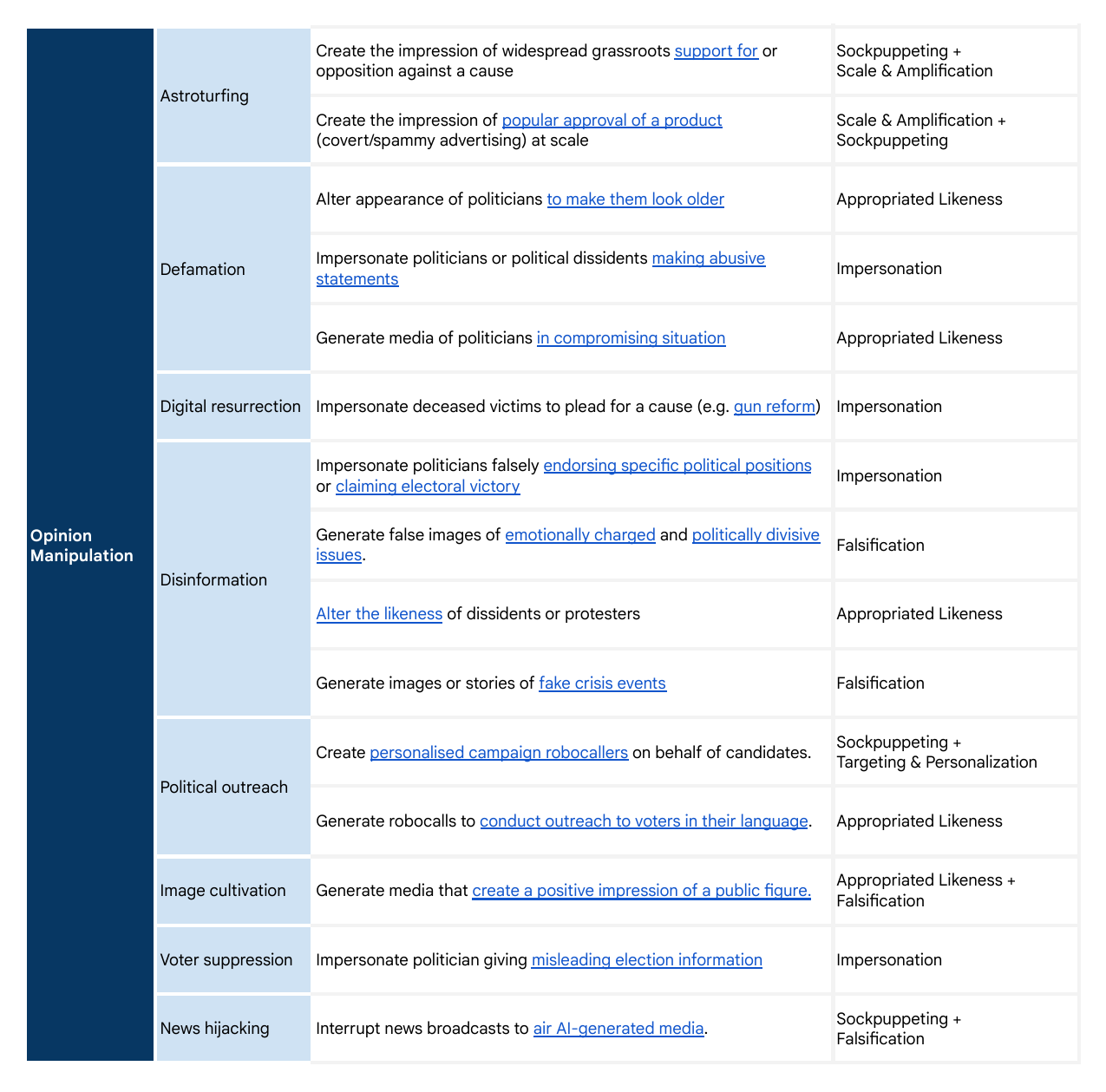}
	\captionsetup{justification=centering}
	\caption*{\textit{(continued on the next page)}}
\end{table}

\clearpage
\setcounter{table}{0}
\begin{table}[ht]
	\centering
	\caption{Strategies of GenAI misuse per goal \textit{(continued)}}
	\includegraphics[width=\columnwidth]{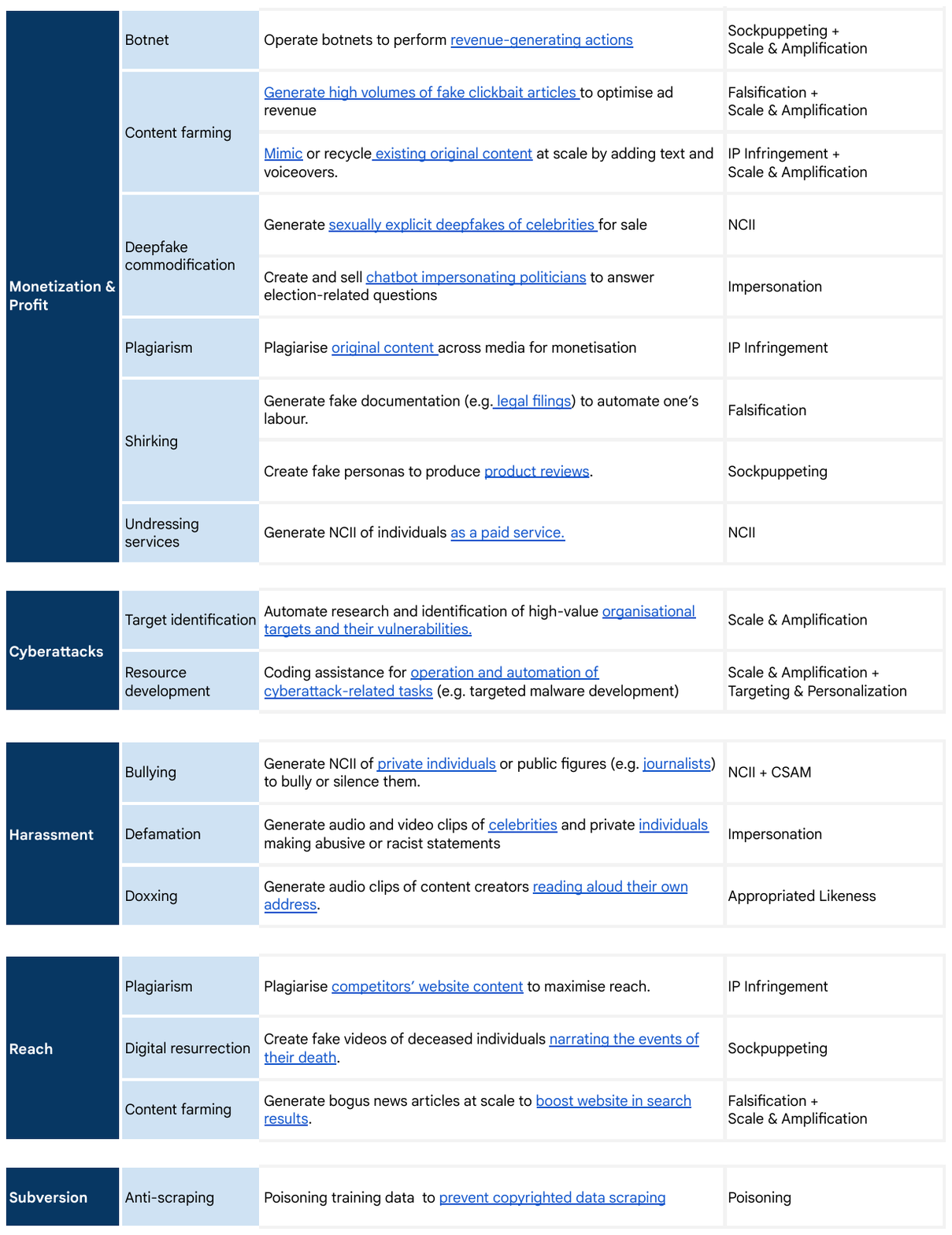}
\end{table}

\end{document}